# Multiobjective Learner Performance-based Behavior Algorithm with Five Multiobjective Real-world Engineering Problems

**Chnoor M. Rahman[1], Tarik A. Rashid[2], Aram Mahmood Ahmed[3, 4] , Seyedali Mirjalili[5,6]**

[1] Applied Computer Department, College of Medicals and Applied Sciences, Charmo University, Sulaimany, Iraq.
[2] Computer Science and Engineering Department, University of Kurdistan Hewler, Erbil, Iraq.
[3] Department of Information Technology, Technical College of Informatics, Sulaimani Polytechnic University, Sulaimany, Iraq.
[4] Department of Information Technology, College of Science and Technology, University of Human Development. Sulaimany, Iraq.
[5]Centre for Artificial Intelligence Research and Optimization, Torrens University, Australia.
[6]Yonsei Frontier Lab, Yonsei University, Seoul, Korea.

## Abstract

In this work, a new multiobjective optimization algorithm called multiobjective learner performance-based behavior algorithm is proposed. The proposed algorithm is based on the process of transferring students from high school to college. The proposed technique produces a set of non-dominated solutions. To judge the ability and efficacy of the proposed multiobjective algorithm, it is evaluated against a group of benchmarks and five real-world engineering optimization problems. Additionally, to evaluate the proposed technique quantitatively, several most widely used metrics are applied. Moreover, the results are confirmed statistically. The proposed work is then compared with three multiobjective algorithms, which are MOWCA, NSGA-II, and MODA. Similar to the proposed technique, the other algorithms in the literature were run against the benchmarks, and the real-world engineering problems utilized in the paper. The algorithms are compared with each other employing descriptive, tabular, and graphical demonstrations. The results proved the ability of the proposed work in providing a set of non-dominated solutions, and that the algorithm outperformed the other participated algorithms in most of the cases.

## 1. Introduction

During the last few decades, some techniques have been proposed to solve engineering optimization problems. The majority of the techniques were based on nonlinear and linear programming methods. They can find the global optimum for simple to medium problems. An immense number of real-world optimization problems have more than one objective function to maximize and/or minimize. These objectives are often conflicting. Biological disciplines, economic, engineering, and automotive problems are examples of such real-world problems. Thus, since the 1970s, researchers have been interested in examining multiobjective optimization algorithms (MOAs); see [1-6]. However, until preparing this paper, no single approach exists to tackle all the various problems with a huge number of conflicting objectives. While we use the MOAs it is difficult to find one utopian solution which satisfies all the conflicting objectives. This is because that utopian solution may satisfy one objective, or a subset of the objectives whilst provides poor regarding another objective, or a subset of objectives. Hence, the MOAs provides a set of tradeoff solutions instead of one utopian solution. The produced set is known as the non-dominated solution set or Pareto optimal solution set. The non-dominated solution set provides solutions with a tradeoff that satisfy all the conflicting objectives without violating any of the existing constraints. The individual solutions in the non-dominated set are called Pareto optimum. All the Pareto optimum solutions are optimal, and compared to the optimum solutions no better solutions exist in the search space when considering all the objectives. None of the Pareto optimum solutions is better than another Pareto optimum solution. However, based on some other information, which is related to the demands of the problem owner, the problem owner can decide which of the Pareto optimum solutions is better. Hence, finding as many Pareto optimum solutions as possible is necessary. Thus, MOAs may produce an infinite number of optimal solutions. Plotting the Pareto optimum solutions produce a Pareto front. The foremost goal of MOAs is to produce a Pareto front for the provided multiobjective optimization problem (MOP) [7, 8].

Optimization algorithms, in general, are utilized to optimize various problems in various fields. For example, in [9], a genetic algorithm (GA) was combined with interior point techniques (IPTs) to optimize a new approach that solves the initial value of the equation of a Painlev´e II, and its variants, utilizing the feed-forward artificial network. The GA was utilized as a global search technique to optimize the weights of the artificial neural network, and the IPTs were utilized as a local search technique. The proposed model was validated using relative investigation with plain Range-Kutta numerical process on nonlinear Painlev´e II through utilizing various magnitudes of forcing factors. The reason for proposing the technique (GA-IPT) was that with enlarging the iterations, the GA's ability to optimize the solutions was reduced, and the improvements in the solutions for a big number of iterations were not reasonable. Hence, the researchers utilized an effective local search technique to reduce the drawbacks of the global search technique for the proposed work. The hybrid method utilized in the work was efficient; however, it was more prone to different runs. Similarly, the study [10] designed a neuro-heuristic schema for the non-linear second order Thomas-Fermi system. To optimize the schema, the GA and sequential quadratic programming were utilized, and it was discovered that the examined schema was feasible, precise, and effective. In [11], a salp swarm algorithm (SSA) with Jaya optimizer was used to optimize the parameters of temperature effect in dam health monitoring utilizing support vector machines. Jaya and SSA were utilized as search engines to select the parameters of least square support vector machines (LSSVM) and support vector machines (SVM). In the proposed work, a hybrid technique was examined to predict the behaviors of concrete dams. The work discussed the influence of incorporating measured temperatures into the model rather than utilizing a model with indirect temperature. The results proved that the proposed model has a small value of prediction errors and residuals.

Evolutionary algorithms (EAs) are excellent examples of optimization algorithms. EAs are stochastic and global search methods. They symbolize the concept of natural selection and the endurance of the fittest. It has been proven that EAs are powerful and robust search techniques [12, 13]. They excel at finding many tradeoff solutions in the sole run. Additionally, due to their ability to explore the large promising region and solve problems with many conflicting objectives, the EAs are well suited to solve MOPs. Hence, one of the effective approaches to solve MOPs is evolutionary multiobjective optimization algorithms (EMOAs). However, EMOAs usually need a large number of objective function evaluations to converge and find the Pareto front [8, 13].

In this work, we introduce some MOAs and propose a novel evolutionary multiobjective optimization algorithm named the multiobjective learner performance-based behavior (MOLPB) algorithm. The basic ideas of this algorithm are based on the process of transferring graduated students from high school to colleges and the behaviors of learners that affect their performance during the college study, and the factors that may help the learners to change their high-school study behaviors that are not effective anymore for studying in the college [14]. The proposed multiobjective algorithm uses the dominance concept to compare the participated objective functions in a specific problem.

The contributions of the algorithm:

- A new evolutionary algorithm is proposed for solving multiobjective optimization problems.
- The main population is divided into some subpopulations depending on the fitness of the individuals.
- Non-dominated approach and crowding distance techniques are used to recognize the better (fitter) solutions.
- The subpopulations direct the algorithm toward a promising area. Because the subpopulation that contains the best individuals has priority to be used for selecting the parents.
- An archive is employed to store the non-dominated solutions to create the Pareto front set.
- Whenever the number of the non-dominated solution in the archive reaches a number bigger than the archive size, then we will apply the crowding distance to the archive to delete the solutions that have a low crowding distance value.
- The results are confirmed by utilizing several benchmarks, including ZDT, and five real-world multiobjective optimization problems.
- The proposed algorithm is compared against NSGA-II, MODA, and MOWCA for optimizing MOP.

# 2. Related Works

In this section, we introduce four multiobjective algorithms. Three of them later tested on the benchmarks and five real-world multiobjective engineering problems. The results of the algorithms are then compared with the proposed algorithm.

## 2.1 Multiobjective Cellular Genetic Algorithm

In [15] a cellular genetic algorithm (cGA) was proposed. The proposed algorithm is called the multiobjective cellular (MOCell) genetic algorithm, which works like a multiobjective version for cGA. cGAs utilize a small neighborhood concept that provides exploration. The individuals here collaborate only with the individuals in the close neighbors. This means that the chosen parents are from close neighbors. Mutation and recombination operators are applied to the chosen individuals to produce offsprings. MOCell utilizes an external archive for storing the observed non-dominated solution in the course of running the algorithm. To provide reasonable diversity in the external archive, the crowding distance examined in NSGA-II was utilized. The crowding distance was also utilized to manage the number of solutions in the external archive. After each iteration, the auxiliary population replaces the old population, and then a feedback technique is executed. During the feedback procedure, some solutions from the external archive are departed to the population, and the same number of randomly chosen solutions from the population will replace them.

## 2.2 Non-dominated Sorting Genetic Algorithm

In [16], an improved version of the non-dominated sorting genetic algorithm (NSGA-II) was proposed. NSGA-II assigned a rank to individual solutions that are equal to the level of nondomination (level 1 is the best; level 2 is the next best, and so on). At first, a random population $P_0$ with size $N$ is produced. Then the genetic operations are used to produce an offspring population $Q_0$ with size $N$. At each generation $(t)$, the two populations $P_t$ and $Q_t$ are combined to produce a merged population $R_t$ with size $2N$. This step makes elitism secure because all the members from the current and previous steps are included in the produced population. The solutions in the new population $R_t$ are then sorted based on the non-dominated solutions. The solutions in the non-dominated set are the best in the combined population. If the size of the non-dominated set is smaller than the population size, all the solutions from the non-dominated set are copied to the new population $(P_{t+1})$. The rest of the members of the $P_{t+1}$ population comes from the next best-non-dominated fronts according to the ranks. This procedure of copying members from the fronts is continued until all the available fronts are finished. The number of members from all the available fronts may be larger than $N$. Hence, the members come from the last front are sorted in descending order, and crowding based comparison is used to choose the best members to fill the newly produced population $(P_{t+1)}$. The genetic operation (selection, crossover, and mutation) are then utilized in the population $(P_{t+1})$ to produce a new offspring population $(Q_{t+1})$.

The crowding distance technique shows the diversity of non-dominated solutions on every side of a specific non-dominated one. The smaller the value produced by the crowding distance shows a better distribution among solutions in a particular region. It directs the process of selection at the different stages of the multiobjective algorithm against a well distributes Pareto optimal front. The crowding distance technique can be utilized in the objective space and the parameter space as well. The $\prec_n$ the operator was used as a crowded comparison operator. For instance, suppose each individual $i$ has two characteristics:

- $i_{rank}$ indicates the rank of nondomination.
- $i_{distance}$ indicates crowding distance.

Hence, $i \prec_n j$ if $i_{rank}$ is smaller than $j_{rank}$ or if the $rank$ of both $i$ and $j$ is equal, but the $distance$ of $i$ is greater than the $distance$ of $j$. This means that between two solutions the one that has a lower rank is better. However, if the ranks are equal the one that locates in a low crowding region is preferred.

Compared to the original NSGA, NSGA-II has several advantages: 1) the computational complexity of the NSGA-II is smaller than the original NSGA. The reason for this is the non-dominated sorting procedure utilized by the NSGA-II. 2) The NSGA-II utilizes the elitism procedure, which according to [17] has a great impact on improving the performance of a genetic algorithm. Moreover, using elitism prevents losing the best-found solution. 3) To protect the diversity of the non-dominated solution set, a crowding-based procedure was utilized which does not need any defined parameters by the user to keep the diversity rate.

## 2.3 Strength Pareto Evolutionary Algorithm

In [18] a new elitist non-dominated multiobjective evolutionary algorithm called Strength Pareto Evolutionary Algorithm (SPEA) was proposed. SPEA has an external population that contains all the non-dominated solutions found so far. The algorithm maintains the external population during all the generations and makes sure that the external population has been involved in all the operations. At the beginning of each generation, a merged population is produced which consists of the current, and the external populations. Depending on the number of solutions dominated by the non-dominated solutions, fitness is given to each non-dominated solution in the merged population. Additionally, fitness worse than the worst fitness of the non-dominated solutions is assigned to the dominated solutions. The procedure of assigning fitness directs the search toward a promising area. Moreover, SPEA uses a deterministic clustering method to provide a good rate of diversity among non-dominated solutions.

## 2.4 Multiobjective Water Cycle Algorithm

A multiobjective water cycle algorithm (MOWCA) is proposed in [19]. The water cycle algorithm (WCA) was first introduced in [20]. WCA imitates the observation of the water cycle and the process of flowing the streams and rivers against the sea. A random population that consists of the raindrops is built as a first step. The best raindrop (individual) represents the sea. The next best raindrops are considered as a river and the remaining raindrops are chosen as streams. The water on the rivers comes from streams. Additionally, the water of the rivers flows to the sea. The amount of water that flows to a river or sea differs from one stream to another.

Similar to nature, here, streams consist of raindrops, and the streams merge to form new rivers. Some of the streams directly flow to the sea. The final point for all streams and rivers is the sea which is considered as the best individual. The procedure of flowing streams in various directions represents the exploitation phase of the algorithm. If the produced solution by a stream is better than the produced solution by a river, the positions of stream and river are swapped. The same procedure of swapping can happen between the sea and river. The condition of evaporation was utilized to avoid trapping into local optima. The sea evaporates whenever a river or a stream flows into it. Hence, the evaporation procedure happens whenever a stream or a river is close enough to the sea. Whenever the evaporation procedure ends, the raining procedure will apply and new streams will generate in various locations. The raining process is the same as the mutation in the genetic algorithm. The exploration phase is secured by the evaporation procedure. Similar to most of the MOAs, MOWCA uses the crowding distance technique to choose the best solutions as rivers and the sea. Moreover, it utilized an external archive to store the non-dominated solutions. Additionally, crowding distance is utilized to control which solution should enter or remain in the external archive when it becomes full. MOWCA has mixed exploration and exploitation. To further clarification, the exploitation phase in the MOWCA comes first, flowing streams against the rivers and rivers against the sea (exploitation phase). In the middle of this moving procedure, evaporation occurs (exploration phase). This technique is significant to explore a wider area of design space while focusing on close optimum non-dominated solutions. For handling constraints, the MOWCA defines a technique. At each generation, when the solution set is defined, the algorithm checks all the constraints, and it separates the feasible solutions. Then, the non-dominated solutions are chosen from the separated feasible solutions. Afterward, these non-dominated solutions are moved to the Pareto archive. Finally, this Pareto archive is used to choose the rivers and sea.

# 3. Learner Performance-Based Behavior Algorithm

In this section, first, the single version of the learner performance-based behavior algorithm is presented, and then the multiobjective version of the algorithm is discussed.

## 3.1 Single-Objective Learner Performance-Based Behavior Algorithm

Learner performance-based behavior (LPB) algorithm inspired by the process of accepting graduate students from high school in colleges. The procedure of transferring high school students to colleges starts with a group of graduate students from high school. Depending on their GPA, some of the applications of these students are accepted in different departments and some of them are rejected. Departments specify the minimum GPA that the students should have. This is like dividing the students into groups depending on their GPA. The students that have a GPA greater than or equal to the minimum required GPA for a specific department are accepted in that department. However, the students with higher GPA have priority to be accepted first. The LPB algorithm utilizes the division probability (*dp*) operator to separate a percentage of individuals (students) randomly [14]. Equation (1) can be used to separate a percentage of individuals from the main population.

$$S = nPop \times dp \tag{1}$$

Where:

S indicates the number of separated individuals from the main population, nPop indicates the number of the main population, *dp* is a value in the range [0.1, 0.9].

After calculating the fitness of individuals in the separated group, the individuals will be divided into two subpopulations (good and bad). A good population contains individuals with better fitness (GPA), and a bad population contains individuals that have lower fitness. After this, the fitness of all individuals in the main population is calculated. In the main population, the individuals that have a fitness smaller than or equal to the best fitness in the bad population go to the bad population, as shown in equation (2). The remaining individuals in the main population are divided into two subpopulations. The individuals that have fitness higher the best fitness in the good population is moved to the perfect population, as shown in equation (3), and those who have fitness smaller than or equal to the best fitness in the good population go to the good population, as shown in equation (4). The individuals in the perfect population have priority to go through the optimization process first, and then the individuals in the good population, and the individuals in the bad population.

$$x \in badPopulation \\ iff\ x \leq \max(badPopulation) \tag{2}$$

$$x \in perfectPopulation \\ iff > \max(goodPopulation) \tag{3}$$

$$x \in goodPopulation \\ iff\ x \leq \max(goodPopulation) \tag{4}$$

Where:

In equations (7-9), the problem is maximization, and $x$ is the fitness of an individual in the main population.

Moreover, fresh students should adopt new studying behaviors to be good college students. Additionally, when students go to colleges, their studying behaviors are affected by the studying behaviors of other students. To show this in the algorithm, a crossover operator was utilized. A crossover operator is utilized to exchange information between two individuals (parents) and it produces two offsprings as a result that have different characteristics.

Moreover, the level of metacognition has a big impact on students. The students that have a good level of metacognition are better compared with those who do not have an adequate level of metacognition. Moreover, when the level of metacognition is affected all the studying behaviors will be affected as well. So that, stochastically exchanging positions of behaviors or updating the values of behaviors according to a specific rate can do that. This was shown in the algorithm by utilizing the mutation operator from the genetic algorithm.

## 3.2 Multiobjective Learner Performance-Based Behavior Algorithm

To change the learner performance-based behavior (LPB) algorithm to an efficacious multiobjective optimization algorithm, we need to redefine the main features of the method (i.e., the learner that owns the best skills). When a single objective function requires to be minimized, the best solutions found so far are chosen as the best learner. However, MOPs have more than one objective to be evaluated (minimized or maximized). Hence, the algorithm should utilize another criterion for selecting the learners (individuals) from the subgroups. Here, crowding distance from [16] is utilized to choose the best non-dominated solutions. As shown in previous sections, the crowding distance procedure shows the diversity of non-dominated solutions on every side of a specific non-dominated one.

The smaller the value produced by the crowding distance shows a better distribution among solutions in a particular region. The crowding distance can be used in both objective and parameter spaces or only in the objective space. To utilize it in the objective space, we sort all the non-dominated solutions according to the result of one of the objectives.

Dividing the main population through utilizing the *dp* operator into various subpopulations and focusing on the subpopulation that has the best individuals so far directs the search toward a promising area. Selecting the individuals from the best subpopulation and then the next best is the best guide for selecting solutions in the next coming iterations. This procedure is counted as an important pace in the multiobjective learner performance-based behavior (MOLPB) algorithm. To provide a good convergence rate and protect a good diversity, the crowding distance technique is applied to all non-dominated solutions in each iteration. Afterward, the subpopulations are rebuilt utilizing the non-dominated solutions that are nominated based on the crowding distance. Therefore, the new subpopulations have individuals with a smaller value of crowding distance. The crowding distance mechanism from NSGA-II is used to maintain a good diversity rate.

Besides, it is crucial to have an archive to store the non-dominated solutions to create the Pareto front set. At each generation, we update the archive and delete the dominated solutions. We assign the archive and the population to the same size. Whenever the number of the non-dominated solution in the archive reaches a number bigger than the archive size, then we will apply the crowding distance to the archive to delete the solutions that have a low crowding distance value.

**Steps of the MOLPB Algorithm**

1. Initialize the operators (*population size, Crossover, Mutation, dp*).
2. Randomly generate the initial population.
3. Randomly choose several individuals (learners) by using the *dp* operator.
4. Run the selected individuals in the previous step through the fitness function.
5. Use the non-dominated approach and crowding distance to choose half of the population in step 3, and call this group goodPopulation. Rename the other half as badPopulation.
6. Do crossover and mutation between half of the elements in the goodPopulation, and for the other half, we bring partners from the main population. However, before choosing the partners from the main population we do the following:
    A. We find the best element (the non-dominated one comparing to the other elements) in the badPopulation.
    B. Remove the elements from the main population that are dominated by the best element founded in A.
7. In the main population, the individuals that have fitness smaller than or equal to the best fitness in the badPopulation go to the badPopulation.
8. The remaining individuals in the main population are divided into two sub populations:
    I. perfectPopulation: contains the individuals that have fitness higher the best fitness in the goodPopulation.
    II. goodPopulation: contains individuals that have fitness smaller than or equal to the best fitness in the goodPopulation.
9. After filtering the data, do crossover and mutation between the rest of the individuals in the goodPopulation and the elements in the perfectPopulation. If the individuals from the perfectPopulation were not enough, bring individuals from the badPopulation.
10. Store the non-dominated solutions in the external archive.
11. Find the crowding distance for the non-dominated solutions in the external archive.
12. Whenever the external archive becomes full, use the crowding distance technique to remove the dominated solutions (those who have low crowding distance value) in the archive and store the non-dominated solutions.
13. If the stopping condition is met, the algorithm will quit, otherwise, go to step 3.

# 4. Metrics

To evaluate the proposed algorithm quantitatively and compare the results with other multiobjective algorithms, four performance metrics were used. The utilized metrics are among the most widely used metrics for evaluating the multiobjective algorithms [21]. These metrics are described in detail in the following subsections.

## 4.1 Generational Distance Metric

Generational Distance (GD) metric was proposed by [22]. According to [21], GD is the second most used metrics by the researchers to examine the MOEAs. It takes the obtained Pareto front (approximation set) and examines how far the set is from the optimal Pareto front. It examines the average distance of Euclidean between the members of the obtained Pareto set and the closest individuals in the optimal Pareto front. This metric is used to count the accuracy of the algorithm to find the Pareto optimal solutions.

## 4.2 Reverse Generational Distance Metric

The Reverse Generational Distance (RGD) is a reverse form of GD; however, the RGD shows remarkable differences compared with the GD. Instead of the average distance of Euclidean, it examines the smallest Euclidean distance among the obtained and optimal Pareto front solutions. Moreover, it utilizes the optimal Pareto front solutions as a reference instead of the solutions in the obtained Pareto set. Additionally, it is utilized to measure the accuracy (convergence) and the diversity of the algorithm [23]. Similar to the GD, RGD is counted as one of the most widely used metrics [21].

## 4.3 Metric of Spacing

The spacing metric (S) is used to measure the distribution of the obtained solutions in the Pareto optimal front. When the spacing value is zero, it means that the space between individuals is similar [24].

## 4.4 Metric of Maximum Spread

The metric of Maximum Spread (MS) is used for examining the diversity of the solutions in the obtained Pareto front. It uses the width sum of every participated objective to show the spread of the solutions [25].

# 5. Results and Discussions

In this section, we tested the proposed multiobjective algorithm using a group of standard benchmarks, and five real-world multiobjective engineering problems. Additionally, to statistically evaluate the ability of the algorithm, the average (Ave.) and standard deviation (Std.) of the results for all the participated algorithms (MOLPB, MOWCA, NSGA-II, and MODA) were calculated by the authors. The results are then compared with three multiobjective algorithms in the literature (MOWCA, NSGA-II, and MODA). The overall parameters for the MOLPB and other participated algorithms are shown in Table 1. The parameters and MATLAB code of MOWCA, NSGA-II, and MODA are from [26], [27], and [28], respectively. However, some of their main parameters are presented in Table 1.

MATLAB programming software was used to code the MOLPB algorithm. For optimizing each benchmark, 30 independent runs were utilized, the algorithm executed over 30 independent runs and 350 iterations each. To store the non-dominated solutions we utilized an external archive with the size of 100. A standard laptop with a processor Intel Core i7, 16 GHz was used. Similarly, the same conditions and computing platform are utilized to run all other participated algorithms (MOWCA, NSGA-II, and MODA) in this paper by the authors of the work.

TABLE 1
PARAMETER SETTINGS FOR Algorithms

| Algorithms | Parameters | Parameter Value |
|---|---|---|
| MOLPB | Crossover rate | 2*round (0.7*population size) |
| | Mutation rate | 0.02 |
| | Population Size | 100 |
| | dp | 0.6 |
| | External Archive Size | 100 |
| MOWCA | Nsr | 4 |
| | dmax | 1e-16 |
| | Population Size | 100 |
| | External Archive Size | 100 |
| NSGA-II | Crossover rate | 2*round (0.7*population size) |
| | Mutation rate | 0.02 |
| | Population Size | 100 |
| | External Archive Size | 100 |
| MODA | Population Size | 100 |
| | External Archive Size | 100 |
| | Archive_X | zeros(100,dim) |
| | Archive_F | ones(100,obj_no)*inf |
| | r | (ub-lb)/2 |
| | V_max | (ub(1)-lb(1))/10 |
| | Food_fitness | inf*ones(1,obj_no) |
| | Food_pos | zeros(dim,1); |
| | Enemy_fitness | -inf*ones(1,obj_no); |
| | Enemy_pos | zeros(dim,1); |

The description of quality in MOPs is significantly more complex compared with that in the optimization of single-objective problems. Hence, reasonable convergence and diversity are used as metrics to recognize better MOAs [21, 29]. To confirm the performance of MOLPB we utilize the ZDT benchmarks.

The description of quality in MOPs is significantly more complex compared to that in the optimization of single-objective problems. Hence, reasonable convergence and diversity are used as metrics to recognize better MOAs [19, 30]. In MOAs, to decide the quality of the algorithm in solving MOPs, researchers focus on the following [24]:

- Minimizing the distance between the non-dominated solutions set in the Pareto optimal front.
- Providing well and uniformly distributed solutions. A distance metric can be used to examine this criterion.

The results of the utilized benchmarks and the real-world engineering problems in this paper proved the perfect distance between the non-dominated solutions produced by the MOLPB algorithm. Additionally, the algorithm can provide a set of the well and uniformly distributed solutions.

Computational complexity is described as a method of size (cardinality) of the input set of data. In general, when debating computational complexity if it is not stated, the time complexity is presumed. For measuring the computational complexity of an algorithm, big-oh notation $(O)$is the most popular complexity measure. It is complicated to measure the time complexity of MOEAs [31]. Because they are stochastic algorithms, the operators they use, the implementation of the operators, the representation of the individuals, the number of objectives, the size of the population, affect their complexity. For calculating the computational complexity of MOEAs, it is important to examine both, the complexity of each generation and the complexity of all generations [31, 32]. Concerning the MOLPB algorithm, the computational complexity for a single generation is $O(n_{ob}{n_{pop}}^2)$, where $n_{ob}$ is the number of objectives, and $n_{pop}$ is the population size. However, for all the generations, the complexity is $O(n_{ob}n_{pop}{n_g}^2)$, where $n_g$ represents the number of generations.

## 5.1 ZDT Benchmarks

The ZDT benchmarks were first proposed in [30]. The family of the ZDT, which consists of six benchmarks, is very popular to assess the performance of the MOPs. Each ZDT benchmark contains a characteristic, which shows a real-world optimization problem that could make it difficult to converge to the Pareto front. All the test functions in the ZDT test suit consist of two objectives. Optimization problems with two objectives are counted as the most popular utilization of multiobjective optimization, mainly within applications in the engineering area [33]. The ZDT5 was not used in this study for benchmarking the algorithm because it is for binary problems. The ZDT functions and their parameters are presented in Table A1 in Appendix A. More information about the ZDT family test suit can be found in [30]. The average and standard deviation for each of the ZDT benchmarks are counted for all the utilized metrics (GD, MS, RGD, and S), and they are shown as (Ave.GD, Ave.MS, Ave.RGD, Ave.S) and (Std.GD, Std.MS, Std.RGD, Std.S), respectively.

The statistical results for the ZDT1 prove that for all the performance metrics the produced results by the MOLPB algorithm are better than the MODA. However, the average values of both GD and RGD of MOWCA and NSGA-II were better than that of the MOLPB algorithm. The values of GD and RGD were evidence that the proposed algorithm is more accurate than MODA, and less accurate in producing results than MOWCA and NSGA-II. On the other hand, the average values of MS and S of the MOLPB algorithm were better than all the participated algorithms. The results of MS and S metrics proved the superior distribution of the solutions produced by the algorithm. The results of the standard deviation proved that the stability of the algorithm compared with the MODA was much better. However, compared to the values of standard deviation, the other two algorithms were more stable. The numerical results of the ZDT1 are shown in Table 2. The non-dominated solutions for all the participated algorithms are shown in Figure 1. As it is shown in Figure 1, the provided Pareto front by the MOLPB algorithm is very close to the optimal one. Regarding the processing time (PT), the MOLPB algorithm is the faster optimizer to optimize this function compared to the other algorithms. The reason for this is that the subpopulations are smaller compared with the main population. Hence, searching for the solutions in these subpopulations is speeder, which minimizes the optimization time.

TABLE 2
COMPARISON OF THE PARTICIPATED ALGORITHMS IN THE LITERATUR BASED ON THE AVERAGE AND STD. VALUES OF THE PERFORMANCE METRICS FOR THE ZDT1

| ZDT1 | Algorithms | | | |
|---|---|---|---|---|
| | MOLPB | MOWCA | NSGA-II | MODA |
| Ave.GD | 0.044265430742418 | **0.002527302780766** | 0.013829780119414 | 1.425031923699603 |
| Ave.MS | **1.064283348714445** | 1.411033597993398 | 1.422797725674114 | 1.615306829628331 |
| Ave.RGD | 0.121369008876936 | **0.007164878638053** | 0.015164718073763 | 1.421448361247811 |
| Ave.S | **0.058113544993055** | 0.077638306434015 | 0.082292770323562 | 0.095326590429206 |
| Std. GD | 0.008864261539852 | 0.004815683618772 | **0.001331554471021** | 0.250049273525705 |
| Std. MS | 0.149466693558889 | 0.012004101170820 | **0.002801744338594** | 0.323842686018692 |
| Std. RGD | 0.046099193565651 | 0.007523316539487 | **0.001295711897951** | 0.157188696968809 |
| Std. S | 0.011140359863496 | 0.011878959230857 | **0.008194862995320** | 0.082947159440500 |
| PT (Seconds) | 417.260033 | 3772.587002 | 36558.762675 | 511.23412 |

The results of the ZDT2 benchmark are presented in Table 3. The average values of the GD and RGD of the NSGA-II were better than the MOLPB algorithm, and the MOLPB was superior to the MODA and MOWCA. This proved the accuracy of the MOLPB algorithm compared with MODA and MOWCA. The average values of the GD and RGD proved the accuracy of the algorithm compared to the MODA and MOWCA. The average results of the MS and S of MODA were superior to the other algorithms. The average results of S of the MOLPB algorithm were comparable to the MOWCA, which proved the diversity of the proposed technique to solve this problem. Additionally, the results of the Std. proved the superior stability of the algorithm. For all the metrics, the Std. of the MOLPB algorithm was the second-lowest among other algorithms. Regarding the PT, the MOLPB algorithm performed much better compared to the other participated algorithms. The produced Pareto front for all the participated algorithms is shown in Figure 2. The formulated Pareto front in Figure 2 shows that the ability of the proposed algorithm is great to optimize the problem, and the provided Pareto front is very similar to the optimal one.

TABLE 3

COMPARISON OF THE PARTICIPATED ALGORITHMS IN THE LITERATUR BASED ON THE AVERAGE AND STD. VALUES OF THE PERFORMANCE METRICS FOR THE ZDT2

| ZDT2 | Algorithms | | | |
|---|---|---|---|---|
| | MOLPB | MOWCA | NSGA-II | MODA |
| Ave.GD | 0.025357244207100 | 0.084791072209500 | **0.012670533233810** | 2.071509410191840 |
| Ave.MS | 1.077046624136208 | 0.797772124718703 | 1.371167131082542 | **0.483434957487415** |
| Ave.RGD | 0.073216310576615 | 0.309530230479231 | **0.015862830875325** | 2.264796385841505 |
| Ave.S | 0.059965108510247 | 0.056601642167611 | 0.080402990827512 | **0.035931906358535** |
| Std. GD | 0.005771383428269 | 0.114259661920143 | **0.001310577589310** | 0.342129006084503 |
| Std. MS | 0.237179797315223 | 0.567637077942365 | **0.073891218290820** | 0.381840838329998 |
| Std. RGD | 0.045832022230383 | 0.240067712168920 | **0.004999387982251** | 0.348115501710479 |
| Std. S | 0.010687808886550 | 0.043451151211900 | **0.009357683244641** | 0.052968724285136 |
| PT (Seconds) | 775.590055 | 25063.840706 | 11187.517668 | 791.561123 |

The results of the performance metrics for ZDT3 are presented in Table 4. The average values of the GD, RGD, and MS for the MOLPB algorithm were much better compared with the MODA, and this proves the accuracy and good divergence of the proposed work compared with the MODA. Additionally, the stability of the algorithm for providing distributed non-dominated solutions was better than the rest of the participated algorithms. This was evaluated by using the standard deviation of the S metric. Moreover, the standard deviation of the MS metric for the MOLPB algorithm was the second lowest value, and this proves the stability of the algorithm in providing accurate non-dominated solutions. However, the values of the metrics show that the proposed algorithm did not perform well on this benchmark compared with the MOWCA and NSGA-II. The reason for this is that for this problem, the distribution of the Pareto optimal solutions produced by those algorithms is better than the distribution of the Pareto optimal solutions produced by the MOLPB algorithm. Similar to the ZDT1 and ZDT2, the processing time of the MOLPB algorithm for solving ZDT3 is smaller than the other participated algorithms. The produced Pareto front for the algorithms is presented in Figure 3. The produced Pareto front is evidence that the algorithm can produce a set of non-dominated solutions which are very close to the optimal solutions.

TABLE 4
COMPARISON OF THE PARTICIPATED ALGORITHMS IN THE LITERATUR BASED ON THE AVERAGE AND STD. VALUES OF THE PERFORMANCE METRICS FOR THE ZDT3

| ZDT3 | Algorithms | | | |
|---|---|---|---|---|
| | MOLPB | MOWCA | NSGA-II | MODA |
| Ave.GD | 0.106004656429132 | **0.006612430617886** | 0.016101442270025 | 1.446026125031364 |
| Ave.MS | 2.043482943305655 | **1.905909113684476** | 1.964728755886758 | 2.239811623045126 |
| Ave.RGD | 0.106791338256898 | **0.013330629334908** | 0.021700855584659 | 1.151288213661917 |
| Ave.S | 0.278531391356936 | 0.236605019426458 | 0.276866700848928 | **0.110423062292349** |
| Std. GD | 0.032140674675092 | 0.012240666949095 | **0.002828703268223** | 0.297496970356768 |
| Std. MS | 0.088450179291428 | **0.087837176771037** | 0.096907980612161 | 0.339411591076640 |
| Std. RGD | 0.019472019350509 | 0.015376573215378 | **0.007008387391151** | 0.192613467802108 |
| Std. S | **0.031100339624892** | 0.055542891565432 | 0.031896496198046 | 0.075200870882026 |
| PT (Seconds) | 3142.86981 | 5594.529833 | 5543.006542 | 3182.895919 |

The statistical results of the ZDT4 are shown in Table 5. In general, the results of the performance metrics prove that the proposed work was the second-best algorithm among the other multiobjective algorithms in the literature. The results are evidence of the accuracy and stability of the algorithm in producing non-dominated solutions. The results of the NSGA-II were better than the produced results by the MOLPB algorithm. On the contrary, the produced results of the MOLPB algorithm were much better compared with the MOWCA and MODA. Hence, the algorithm could prove its ability in optimizing the problem and providing well-distributed solutions with great accuracy. Figure 4 shows the convergence curve of the MOLPB algorithm, MOWCA, NSGA-II, and MODA, for optimizing the ZDT4 test function. Figure 4 proved the truth that the MOLPB algorithm is the second best algorithm to optimize this problem. The produced Pareto front by the MOLPB algorithm was very close to the optimal Pareto front, which is evidence that the multiobjective version of the proposed algorithm can optimize this problem accurately, produce solutions that are well distributed, and own a considerable diversity. Moreover, the examined algorithm converged earlier compared with the MOWCA and NSGA-II. However, the PT for the MODA is smaller compared with the MOLPB.

TABLE 5
COMPARISON OF THE PARTICIPATED ALGORITHMS IN THE LITERATUR BASED ON THE AVERAGE AND STD. VALUES OF THE PERFORMANCE METRICS FOR THE ZDT4

| ZDT4 | Algorithms | | | |
|---|---|---|---|---|
| | MOLPB | MOWCA | NSGA-II | MODA |
| Ave.GD | 0.456998671317749 | 20.555347065303053 | **0.065371555050419** | 24.379445817448893 |
| Ave.MS | 2.658531869844119 | 37.214693559291180 | **1.369420372387415** | 15.986878682236876 |
| Ave.RGD | 0.477817266466399 | 0.312606141602881 | **0.073018912034728** | 17.397825417742403 |
| Ave.S | 0.269385761145204 | 5.447213583367873 | **0.083708788544184** | 3.772289949886725 |
| | | | | |
| Std. GD | 0.586633391348093 | 19.881301999388040 | **0.090770978045562** | 8.835654194500952 |
| Std. MS | 5.119876239461383 | 29.104464338945050 | **0.127984126339398** | 11.550139937919003 |
| Std. RGD | 0.209062332124054 | 0.394787901157078 | **0.096862393148957** | 7.289940777779140 |
| Std. S | 0.624637206705755 | 5.306785619978045 | **0.024756837408573** | 3.401976460838918 |
| | | | | |
| PT (Seconds) | 4435.67955 | 20522.923772 | 6044.928297 | 1405.00395 |

Similar to ZDT4, the MOLPB algorithm has a great ability to optimize the ZDT6 test function. The results of this problem prove the superiority of the algorithm to provide a set of Pareto fronts which is very similar to the optimal Pareto front. The accuracy of the algorithm in providing optimal solutions is proven by the value of the GD metric. Moreover, the considerable diversity of the algorithm is shown through the result of the MS metric. The results of the RGD and S metrics of the NSGA-II were better than the results of the MOLPB algorithm. On the contrary, the proposed work produced better results compared with the MOWCA and MODA in all the metrics, and in the GD and MS metrics compared with the NSGA-II. The statistical results of the ZDT6 are provided in Table 6. Figure 5 shows the convergence curve for ZDT6 by the MOLPB algorithm, MOWCA, NSGA-II, and MODA, respectively. Figure 5 provides evidence of that presented in Table 6. The PT in Table 6 shows that the proposed algorithm is the fastest compared with the other algorithms.

TABLE 6
COMPARISON OF THE PARTICIPATED ALGORITHMS IN THE LITERATUR BASED ON THE AVERAGE AND STD. VALUES OF THE PERFORMANCE METRICS FOR THE ZDT6

| ZDT6 | Algorithms | | | |
|---|---|---|---|---|
| | MOLPB | MOWCA | NSGA-II | MODA |
| Ave.GD | 0.035370731462601 | 0.472209953362659 | **3.724934272519937e-04** | 3.697731156014888 |
| Ave.MS | 1.421201407339415 | 4.021422537734907 | **1.168687086091388** | 2.495071009099371 |
| Ave.RGD | 0.035066129960508 | 0.537197936559747 | **0.004126324713478** | 3.327913863373460 |
| Ave.S | 0.109923088428612 | 0.317449425745520 | **0.083879755246637** | 0.339247135308378 |
| | | | | |
| Std. GD | 0.107416279119494 | 0.580964481089794 | **1.138339682725477e-04** | 2.009018348088154 |
| Std. MS | 0.963582519420238 | 3.025667272224291 | **4.790800770946798e-16** | 1.768359843926890 |
| Std. RGD | 0.061738082413927 | 0.209160533380801 | **2.155734266302482e-04** | 2.060644809057377 |
| Std. S | 0.098539051341019 | 0.291789426914573 | **0.008136504784781** | 0.416668639996970 |
| | | | | |
| PT (Seconds) | 2892.86229 | 3820.937869 | 5667.043311 | 38004.315963 |

## 5.2 Real-World Engineering Problems

In this section, the proposed algorithm was tested on five multiobjective real-world engineering problems. All the problems were tested on several other algorithms in [34], and the optimal Pareto front for all the problems is presented. The optimal Pareto front in [34] was utilized to calculate the metrics (GD and RGD), and visualize the obtained Pareto front with the optimal one whenever needed. The problems are described in the following subsections. The average and standard deviation for each of all the problems is counted for the utilized metrics (GD, and RGD), and they are shown as (Ave.GD, and Ave.RGD) and (Std.GD, and Std.RGD), respectively.

### 5.2.1 The Four Bar Truss Design Problem

The four-bar truss design problem is one of the most popular problems for evaluating and validating various techniques [19]. This problem is also utilized in [35, 36]. The mathematical form of the problem is as follows:

Minimize

$$f_1(x) = L(2x_1 + \sqrt{2}x_2 + \sqrt{x_3} + x_4) \quad (5)$$

$$f_2(x) = \frac{FL}{E}\left(\frac{2}{x_1} + \frac{2\sqrt{2}}{x_2} - \frac{2\sqrt{2}}{x_3} + \frac{2}{x_4}\right) \quad (6)$$

Where

$$\left(\frac{F}{\sigma}\right) \le x_1, x_4 \le 3 \times \left(\frac{F}{\sigma}\right)$$

$$\sqrt{2} \times \left(\frac{F}{\sigma}\right) \le x_2, x_3 \le 3 \times \left(\frac{F}{\sigma}\right)$$

The parameters are as follows:

F = 10 kN, E = $(2)10^5$ kN/cm$^2$ , L = 200 cm, $\sigma$ = 10 kN/cm$^3$ .

Shifting the joint and the magnitude of the four-bar truss should be optimized at the same time. The results of various techniques for this problem are shown in Table 7.

TABLE 7
COMPARISON OF THE PARTICIPATED ALGORITHMS IN THE LITERATUR BASED ON THE AVERAGE AND STD. VALUES OF THE PERFORMANCE METRICS FOR THE FOUR BAR TRUSS DESIGN PROBLEM

| Algorithms | GD | | RGD | |
|---|---|---|---|---|
| | Ave. | Std. | Ave. | Std. |
| MOLPB | 7.356187147729127 | 0.10024549794804336e+2 | 0.28246974355565293e+2 | 0.13866918530380362e+2 |
| MOWCA | 1.118324970001251e+02 | 0.33800859012025320e+2 | 1.344468357259677e+02 | 1.052978069707367e+02 |
| NSGA-II | **1.892380652813678** | **6.194958934979878** | 1.115290579364851e+02 | 0.25798795805055256e+2 |
| MODA | 0.23634177242407645e+2 | 0.16853528765017117e+2 | **0.21102208608179346e+2** | **7.085411854194526** |

Depending on the average of the GD values presented in Table 7, the NSGA-II owned the best performance to solve this problem, and the MOLPB algorithm has the second-best performance for finding the non-dominated solution among the participated algorithms. This proves that the produced non-dominated solutions by the MOLPB algorithm have the smallest distance to the optimal Pareto front after the NSGA-II. However, the average value of the RGD of the MOLPB algorithm was smaller than the RGD value of the NSGA-II. Based on the results of the RGD method, the MOLPB has the second-best diversity among its non-dominated solutions compared to the other algorithms and the MODA has the best diversity. Besides, the value of the Std. for the MOLPB algorithm for both methods (GD and RGD) is the second-best, which proves its stability regarding the non-dominated solutions that have the lowest distance from the optimal Pareto front, and the solutions that have a good diversity in the non-dominated set. The produced Pareto front by the participated algorithms for this problem is shown in Figure 6. The formulated Pareto front proved the provided data in Table 7, and that the algorithm has a great ability to optimize the problem.

### 5.2.2 The Pressure Vessel Design Problem

Both ends of the cylindrical pressure vessel are covered with half-round heads as shown in Figure 7. For this problem, the costs of forming, material, and welding should be minimized, and this is the first objective of the problem. The decision variables are the width of the shell, the width of the head, the radius, and the cylindrical length. $x_1, x_2, x_3$, and $x_4$ are utilized to represent the decision variables, respectively [34, 37].

The mathematical form of this problem is as follows.

Minimize

$$f_1(x) = 0.6224x_1x_3x_4 + 1.7781x_2x_3^2 + 3.1661x_1^2x_4 + 19.84x_1^2x_3 \quad (7)$$

Subject to

$$g_1(x) = x_1 - 0.0193x_3 \geq 0 \quad (8)$$

$$g_2(x) = x_2 - 0.00954x_3 \geq 0 \quad (9)$$

$$g_3(x) = \pi x_3^2x_4 + \frac{4}{3}\pi x_3^3 - 1296000 \geq 0 \quad (10)$$

Where
$x_1, x_2 \in \{1, \ldots\ldots, 100\}$, $x_3 \in [10, 200]$, $x_4 \in [10, 240]$.

The second objective for the mentioned problems is the summation of the constraints. Following is the mathematical formulation of this objective:

$$f_2(x) = \sum_{i=1}^{3} max\{g_i(x), 0\} \quad (11)$$

TABLE 8
COMPARISON OF THE PARTICIPATED ALGORITHMS IN THE LITERATUR BASED ON THE AVERAGE AND STD. VALUES OF THE PERFORMANCE METRICS FOR THE PRESSURE VESSEL DESIGN PROBLEM

| Algorithms | GD | | RGD | |
|---|---|---|---|---|
| | Ave. | Std. | Ave. | Std. |
| MOLPB | **5.229069244360944e+02** | 0.76606682134282340e+02 | **1.289388701987478e+04** | 6.109367308703447e+03 |
| MOWCA | 1.916554037154009e+03 | 3.420966646903447e+03 | 2.129006430224141e+05 | 2.137883752394241e+05 |
| NSGA-II | 5.405377164921749e+02 | **0.3665932587233484e+02** | 4.815574173930896e+04 | 4.125645271030495e+04 |
| MODA | 7.979135951624052e+02 | 9.107417632226021e+02 | 1.885567165197396e+04 | **2.754524212006902e+03** |

For this problem, the MOLPB algorithm produced the lowest average value of GD compared with the other algorithms. This proves the superior performance of the MOLPB algorithm to find the non-dominated solutions with the minimum distance from the optimal Pareto front. Additionally, the average value of the RGD again proved the superior diversity of the MOLPB algorithm in comparison with the other participated algorithms. However, based on the values of the Std., the MOLPB algorithm is the second-best concerning the stability of finding the non-dominated solutions with the minimum distance from the optimal Pareto front and the non-dominated solutions that own a good diversity. As the result of GD proved, the NSGA-II owns the best stability in finding the non-dominated solutions with the minimum distance from the optimal Pareto front. According to the results of RGD, the MODA owns the best stable diversity among the non-dominated solutions. Figure 8 presents a comparison between the MOLPB algorithm, NSGA-II, MOWCA, and MODA. The results from Table 8 are verified in Figure 8.

### 5.2.3 The Coil Compression Spring Design Problem

This problem is a real-world mechanical engineering optimization problem. The spring is a spiral squeezing spring as shown in Figure 9. A strictly lengthwise and a sustained load are put in the spring. Minimizing the volume of steel wire was utilized to create the spring. The decision variables include the number of spirals of the spring ($x_1$), the exterior diameter of the spring ($x_2$), and the diameter of the spring wire ($x_3$). This problem contains integer ($x_1$), continuous ($x_2$), and discrete ($x_3$) variables. The allowable discrete values of $x_3$ are shown in Table 9. This problem contains two objectives to minimize and six constraints. Minimizing the volume of steel wire utilized to create the spring is the first objective. The other objective is the total of the constraint violations [34, 38].

TABLE 9
ALLOWABLE DIAMETERS FOR THE SPRING WIRE

| Allowable diameters for the spring wire | | | | | |
|---|---|---|---|---|---|
| 0.009 | 0.0095 | 0.0104 | 0.0118 | 0.0128 | 0.0132 |
| 0.014 | 0.015 | 0.0162 | 0.0173 | 0.018 | 0.02 |
| 0.023 | 0.025 | 0.028 | 0.032 | 0.035 | 0.041 |
| 0.047 | 0.054 | 0.063 | 0.072 | 0.08 | 0.092 |
| 0.105 | 0.12 | 0.135 | 0.148 | 0.162 | 0.177 |
| 0.192 | 0.207 | 0.225 | 0.244 | 0.263 | 0.283 |
| 0.307 | 0.331 | 0.362 | 0.394 | 0.4375 | 0.5 |

This problem mathematically formulated as follows:

$$f_1(x) = \frac{\pi^2 x_2 x_3^2 (x_1 + 2)}{4} \tag{12}$$

Subject to

$$g_1(x) = -\frac{8C_f F_{max} x_2}{\pi x_3^3} + S \geq 0 \tag{13}$$

$$g_2(x) = -l_f + l_{max} \geq 0 \tag{14}$$

$$g_3(x) = -3 + \frac{x_2}{x_3} \geq 0 \tag{15}$$

$$g_4(x) = -\sigma_p + \sigma_{pm} \geq 0 \tag{16}$$

$$g_5(x) = -\sigma_p - \frac{F_{max} - F_p}{K} - 1.05(x_1 + 2)x_3 + l_f \geq 0 \tag{17}$$

$$g_6(x) = -\sigma_w + \frac{F_{max} - F_p}{K} \geq 0 \tag{18}$$

$$C_f = \frac{4\left({x_2}/{x_3}\right) - 1}{4\left({x_2}/{x_3}\right) - 4} + \frac{0.615 x_3}{x_2} \tag{19}$$

$$K = \frac{G x_3^4}{8 x_1 x_2^3} \tag{20}$$

$$\sigma_p = \frac{F_p}{K} \tag{21}$$

$$l_F = \frac{F_{max}}{K} + 1.05(x_1 + 2)x_3 \tag{22}$$

Where
$x_1 \in \{1, \ldots\ldots, 70\}$
$x_2 \in [0.60, 30]$
$x_3$ is the diameter of the wire and it is shown in Table 9.

And the parameters are as follows:
$F_{max} = 1000\ lb$, it is the highest working load
$S = 189{,}000\ psi$, it is the accepted highest sheer stress
$l_{max} = 14\ inch$, it is the highest free length
$d_{min} = 0.2\ inch$, it is the lowest diameter of the wire
$D_{max} = 3\ inch$, it is the highest exterior diameter of the spring
$F_p = 300\ lb$, it is the preload compression force
$\sigma_{pm} = 6\ inch$, it is the accepted highest diversion under preload
$\sigma_w = 1.25\ inch$, it is the diversion from preload location to highest load location
$G = 11.5 \times 10^6$, it is the material's shear modulus.

$$f_2(x) = \sum_{i=1}^{6} max\{g_i(x), 0\} \tag{23}$$

The produced results in Table 10 show that the MOLPB algorithm outperforms other participated algorithms to solve the coil compression spring design problem. The proposed algorithm produced the minimum average result for both the GD and RGD methods. These produced results for the GD technique is evidence for the superior performance of the MOLPB algorithm to find the non-dominated solutions with the smallest distance from the optimal Pareto front. Moreover, the lowest average result for the RGD proved the superior diversity of the proposed algorithm. Additionally, the minimum result of the std. for the MOLPB algorithm compared with the other algorithms showed the stability of the algorithm. Figure 10 proved the results shown in Table 10.

TABLE 10
COMPARISON OF THE PARTICIPATED ALGORITHMS IN THE LITERATUR BASED ON THE AVERAGE AND STD. VALUES OF THE PERFORMANCE METRICS FOR THE COIL COMPRESSION SPRING DESIGN PROBLEM

| Algorithms | GD | | RGD | |
|---|---|---|---|---|
| | Ave. | Std. | Ave. | Std. |
| MOLPB | 0.001165137348271 | 5.953683716862298e-04 | 3.907638898625130e+02 | 6.995757373604563e+02 |
| MOWCA | 0.036219782034087 | 0.162110588997669 | 4.782498529340439e+04 | 1.086536014708315e+05 |
| NSGA-II | 1.766560342824665e+02 | 6.683175803097043e+02 | 3.579777017698174e+03 | 6.631512913748524e+03 |
| MODA | 1.468585635356069e+04 | 2.764022957116235e+04 | 3.579274283635858e+03 | 2.297668234226298e+03 |

### 5.2.4 The Speed Reducer Design Problem

This problem was first designed as a single objective optimization problem. It was then converted to many-objective optimization. It is the design of a gearbox that can be utilized in some airplanes. It contains three objectives, eleven constraints, and seven decision variables. The first two objectives of the problem are to reduce the volume and the stress in either of the gear shafts, respectively. The third objective is the total of the constraints. This problem is also utilized to test other algorithms in [34, 36, 39]. The mathematical formulation of the problem is as follows:

$$f_1(x) = 0.7854x_1x_2^2\left(\frac{10x_3^2}{3} + 14.933x_3 - 43.0934\right) -1.508x_1(x_6^2 + x_7^2) + 7.477(x_6^3 + x_7^3) +0.7854(x_4x_6^2 + x_5x_7^2) \tag{24}$$

$$f_2(x) = \frac{\sqrt{\left(745x_4/_{x_2x_3}\right)^2 + 1.69 \times 10^7}}{0.1x_6^3} \tag{25}$$

$$g_1(x) = \frac{1}{27} - \frac{1}{x_1x_2^3x_3} \geq 0 \tag{26}$$

$$g_2(x) = \frac{1}{397.5} - \frac{1}{x_1x_2^2x_3^2} \geq 0 \tag{27}$$

$$g_3(x) = \frac{1}{1.92} - \frac{x_4^3}{x_2x_3x_6^4} \geq 0 \tag{28}$$

$$g_4(x) = \frac{1}{1.93} - \frac{x_5^3}{x_2x_3x_7^4} \geq 0 \tag{29}$$

$$g_5(x) = 40 - x_2x_3 \geq 0 \tag{30}$$

$$g_6(x) = 12 - \frac{x_1}{x_2} \geq 0 \tag{31}$$

$$g_7(x) = -5 + \frac{x_1}{x_2} \geq 0 \tag{32}$$

$$g_8(x) = -1.9 + x_4 - 1.6x_6 \geq 0 \tag{33}$$

$$g_9(x) = -1.9 + x_5 - 1.1x_7 \geq 0 \tag{34}$$

$$g_{10}(x) = 1300 - f_2(x) \geq 0 \tag{35}$$

$$g_{11}(x) = 1100 - \frac{\sqrt{\left({745x_5}/{x_2x_3}\right)^2 + 1.575 \times 10^8}}{0.1x_7^3} \geq 0 \tag{36}$$

Where
$x_1 \in [2.6, 3.6]$, it indicates the width of the gear face. $x_2 \in [0.7, 0.8]$, it is the module of the teeth. $x_3 \in \{17, \dots, 28\}$, it indicates the number of teeth in the gear. $x_4 \in [7.8, 8.3]$, it is the space among the bearings on the first cylinder. $x_5 \in [7.8, 8.3]$, it is the space among the bearings on the second cylinder. $x_6 \in [2.9, 3.9]$, and $x_7 \in [5, 5.5]$, are the diameters of the first and second cylinders, respectively.

The third objective is the total of all the constraints, as follows:

$$f_3 = \sum_{i=1}^{11} max\{g_i(x), 0\} \tag{37}$$

In general, the results in Table 11 show that MOLPB was the second-best algorithm to solve this problem. The NSGA-II produced the smallest value for the GD method. Comparing to the rest of the algorithms in this work, the MOLPB algorithm produced the minimum average value of the GD technique. This proves the superiority of both NSGA-II and MOLPB to find the non-dominated solutions with the smallest distance from the optimal Pareto front. Additionally, the superiority of the proposed algorithm was proved for finding non-dominated solutions that have good diversity. The result of the RGD method is reasonable evidence for the superior diversity of the algorithm. Moreover, the results of the Std. showed the stability of the MOLPB algorithm to solve the problem. Figure 11 shows the produced Pareto front by the participated algorithms. The formulated Pareto front is evident to that provided in Table 11.

TABLE 11
COMPARISON OF THE PARTICIPATED ALGORITHMS IN THE LITERATUR BASED ON THE AVERAGE AND STD. VALUES OF THE PERFORMANCE METRICS FOR THE SPEED REDUCER DESIGN PROBLEM

| Algorithms | GD | | RGD | |
|---|---|---|---|---|
| | Ave. | Std. | Ave. | Std. |
| MOLPB | 0.11300957429052747e+02 | 5.353901079223076 | 4.417567589123142e+02 | 1.204597287830257e+02 |
| MOWCA | 0.12093211543779878e+02 | 8.095693487854756 | 4.706728443589605e+02 | 3.351003926974380e+02 |
| NSGA-II | 9.248983587521442 | 4.355702417490958 | 8.757171216956020e+02 | 1.433600470583269e+02 |
| MODA | 0.48368624551698005e+02 | 0.27723667813065873e+02 | 1.259847489533345e+02 | 0.757966829513193e+02 |

## 5.2.5 The Car Side Impact Design Problem

This problem was also addressed in [34, 40, 41]. It consists of four objectives. The aim of the first three objectives ($f_1, f_2$, and $f_3$) is reduce the heaviness of the car, the pubic force that passengers went through, and the V-pillar's total velocity which is in charge of resisting the impact load, respectively. The original problem consists of eleven decision variables. However, since we use the problem from [34], the four stochastic variables are excluded. The formulation of the problem is as follows:

$$f_1(x) = 1.98 + 4.9x_1 + 6.67x_2 + 6.98x_3 + 4.01x_4 + 1.78x_5 + 10^{-5}x_6 + 2.73x_7 \tag{38}$$

$$f_2(x) = 4.72 - 0.5x_4 - 0.19x_2x_3 \tag{39}$$

$$f_3(x) = 0.5\left(V_{MBP}(x) + V_{FD}(x)\right) \tag{40}$$

$$g_1(x) = 1 - 1.16 + 0.3717x_2x_4 + 0.0092928x_3 \geq 0 \tag{41}$$

$$g_2(x) = 0.32 - 0.261 + 0.0159x_1x_2 + 0.06486x_1 \\ +0.019x_2x_7 - 0.0144x_3x_5 - 0.0154464x_6 \geq 0 \quad (42)$$

$$g_3(x) = 0.32 - 0.214 - 0.00817x_5 + 0.045195x_1 \\ +0.0135168x_1 - 0.03099x_2x_6 + 0.018x_2x_7 - 0.007176x_3 \\ -0.023232x_3 + 0.00364x_5x_6 + 0.018x_2^2 \geq 0 \quad (43)$$

$$g_4(x) = 0.32 - 0.74 + 0.61x_2 + 0.031296x_3 + 0.031872x_7 - 0.227x_2^2 \geq 0 \quad (44)$$

$$g_5(x) = 32 - 28.98 - 3.818x_3 + 4.2x_1x_2 - 1.27296x_6 + 2.68065x_7 \geq 0 \quad (45)$$

$$g_6(x) = 32 - 33.86 - 2.95x_3 + 5.057x_1x_2 + 3.795x_2 + 3.4431x_7 - 1.45728 \geq 0 \quad (46)$$

$$g_7(x) = 32 - 46.36 + 9.9x_2 + 4.4505x_1 \geq 0 \quad (47)$$

$$g_8(x) = 4 - f_2(x) \geq 0 \quad (48)$$

$$g_9(x) = 9.9 - V_{MBP}(x) \geq 0 \quad (49)$$

$$V_{MBP}(x) = 10.58 - 0.674x_1 - 0.67275x_2 \quad (50)$$

$$V_{FD}(x) = 16.45 - 0.489x_3x_7 - 0.843x_5x_6 \quad (51)$$

Where

$x_1 \in [0.5, 1.5]$ indicates the width of B-Pillar inner, $x_2 \in [0.45, 1.35]$ indicates the width of B-Pillar augmentation, $x_3 \in [0.5, 1.5]$ indicates the width of the interior floor side, $x_4 \in [0.5, 1.5]$ indicates the width of cross members, $x_5 \in [0.875, 2.625]$ indicates the width of the door beam, $x_6 \in [0.4, 1.2]$ indicates the width of door augmentation, $x_7 \in [0.4, 1.2]$ indicates the width roof rail.

Objective number four is the total of the constraints, as follows:

$$f_4(x) = \sum_{i=1}^{10} max\{g_i(x), 0\} \quad (52)$$

As shown in Table 12, the results show the ability of the MOLPB algorithm to solve this problem. In addition to the high number of objective, decision variables, and constraints the proposed algorithm performed better than the other algorithms to solve the problem. The results proved the super-diversity and performance of the algorithm. It performed superior to produce non-dominated solutions with a small distance from the optimal Pareto front. Additionally, it owned a good diversity and this was proved by the results of the RGD method. The small value of the Std. showed the amazing stability of the algorithm. Moreover, the result of the GD of the MOLPB algorithm was better compared with the other algorithms, and this proves the accuracy of the produced results by the algorithm. Figure 12 shows the produced Pareto front by the participated algorithms. As shown, the proposed algorithm provided a better Pareto front compared with other algorithms. This is evidence of the ability of the proposed algorithm to optimize multiobjective problems and produce accurate results.

TABLE 12
COMPARISON OF THE PARTICIPATED ALGORITHMS IN THE LITERATUR BASED ON THE AVERAGE AND STD. VALUES OF THE PERFORMANCE METRICS FOR THE CAR SIDE-IMPACT DESIGN PROBLEM

| Algorithms | GD | | RGD | |
|---|---|---|---|---|
| | Ave. | Std. | Ave. | Std. |
| MOLPB | 0.303015888766268 | 0.030579816675289 | 1.567421887332045 | 0.285633108011381 |
| MOWCA | 3.927957536364954 | 0.359717355795984 | 8.715478788882232 | 0.345128037040276 |
| NSGA-II | 0.1993309223566308e+02 | 0.875884770917414 | 9.587727366029336 | 0.184723790431805 |
| MODA | 0.310141340133575 | 0.078033566503663 | 2.070993601511424 | 0.623158723216559 |

# 6. Conclusions

In this work, a new multiobjective algorithm called MOLPB was proposed. The MOLPB algorithm is a brand new algorithm for solving problems that have many objectives. The basic ideas of the MOLPB algorithm were motivated by the process of transferring graduated learners from high school to university and improving the studying behaviors of the learners at colleges. The proposed multiobjective algorithm was utilized to optimize a group of benchmarks, and five real-world engineering problems. To confirm the efficacy and ability of the algorithm several criteria were utilized. The produced statistical results from the metrics are evidence that the algorithm can approach the optimal Pareto front and produce a set of good non-dominated solutions compared with other multiobjective algorithms in the literature. Depending on the reported results, the MOLPB algorithm, in general, provides better or competitive results compared with the other multiobjective algorithms. However, in some cases, the NSGA-II outperformed the proposed algorithm. However, the diversity of the algorithm almost was better than the rest of the algorithms in the literature. Although, the MOLPB algorithm proved its ability to optimize different real-world engineering problems and a group of benchmarks, the nature of the problem may affect the performance of the algorithm similar to any other algorithms. In general, the proposed multiobjective algorithm is a suitable technique to provide Pareto optimal solutions for various multiobjective optimization problems.

For future works, several research directions can be recommended. Firstly, the authors recommend utilizing the proposed technique to optimize different problems and compare the results with other multiobjective heuristic techniques. Additionally, proposing a new technique to utilize in the MOLPB algorithm to recognize non-dominated solutions instead of the crowding-based technique is highly recommended.

# Appendices

## Appendix A

TABLE A1
ZDT BENCHMARKS AND THEIR PARAMETERS [25]

| Function Name | Objective Functions | Ranges | Design Variables (n) |
|---|---|---|---|
| ZDT1 | $f_1(x) = x_1$<br>$f_2(x) = g(x)\left[1 - \sqrt{x_1/g(x)}\right]$<br>g(x) is defined as:<br>$g(x) = 1 + 9\left(\sum_{i=2}^{n} x_i\right)/(n-1)$ | [0, 1] | 30 |
| ZDT2 | $f_1(x) = x_1$<br>$f_2(x) = g(x)\left[1 - \left(\frac{x_1}{g(x)}\right)^2\right]$<br>g(x) is defined as:<br>$g(x) = 1 + 9\left(\sum_{i=2}^{n} x_i\right)/(n-1)$ | [0, 1] | 30 |
| ZDT3 | $f_1(x) = x_1$<br>$f_2(x) = g(x)\left[1 - \sqrt{\frac{x_1}{g(x)}} - \frac{x_1}{g(x)}\sin(10\pi x_1)\right]$<br>g(x) is defined as:<br>$g(x) = 1 + 9\left(\sum_{i=2}^{n} x_i\right)/(n-1)$ | [0, 1] | 30 |
| ZDT4 | $f_1(x) = x_1$<br>$f_2(x) = g(x)\left[1 - \sqrt{\frac{x_1}{g(x)}}\right]$<br>g(x) is defined as:<br>$g(x) = 1 + 10(n-1) + \sum_{i=2}^{n}[x_i^2 - 10\cos(4\pi x_i)]$ | $x_1 \in [0,1]$<br>$x_2, \ldots, x_n \in [-5,5]$ | 10 |
| ZDT6 | $f_1(x) = 1 - 1 - e^{(-4x_1)}\sin^6(6\pi x_1)$<br>$f_2(x) = g(x)\left[1 - \left(\frac{f_1(x)}{g(x)}\right)^2\right]$<br>g(x) is defined as:<br>$g(x) = 1 + 9\left[\frac{\sum_{i=2}^{n} x_i}{(n-1)}\right]^{\frac{1}{4}}$ | [0, 1] | 10 |